%% file: main.tex
\documentclass[twocolumn]{article} 

\usepackage{multicol}  
\usepackage[letterpaper, top=0.6in, bottom=0.5in, inner=0.75in, outer=0.75in, headsep=0.3in, footskip=0.3in]{geometry}

\usepackage{titlesec}
\titleformat{\section}{\large\bfseries}{\thesection}{1em}{}
\titleformat{\subsection}{\normalsize\bfseries}{\thesubsection}{1em}{}

\usepackage{graphicx}         
\usepackage{amsmath, amssymb} 
\usepackage{hyperref}          
\usepackage{natbib}          
\usepackage{geometry}         
\usepackage{listings}
\usepackage{pgfplots}
\usepackage{amssymb} 
\usepackage{pifont}  
\usepgfplotslibrary{fillbetween}
\usepackage{algorithm}
\usepackage{algorithmic}
\usepackage{amsmath} 
\usepackage{appendix}

\title{\textbf{AlphaMaze: Enhancing Large Language Models' Spatial Intelligence via GRPO}}

\author{
    Alan Dao (Gia Tuan Dao)\textsuperscript{1}, Dinh Bach Vu\textsuperscript{1} \\
    Menlo Research \\
    \texttt{alan@menlo.ai, bach@menlo.ai} \\
    \textsuperscript{1}Equal contribution.
}

\date{February 20, 2025} 

\begin{document}

\maketitle
\begin{abstract}
\noindent Large Language Models (LLMs) have demonstrated impressive capabilities in language processing, yet they often struggle with tasks requiring genuine visual spatial reasoning. In this paper, we introduce a novel two-stage training framework designed to equip standard LLMs with visual reasoning abilities for maze navigation. First, we leverage Supervised Fine-Tuning (SFT) on a curated dataset of tokenized maze representations to teach the model to predict step-by-step movement commands. Next, we apply Group Relative Policy Optimization (GRPO)—a technique used in DeepSeek-R1—with a carefully crafted reward function to refine the model’s sequential decision-making and encourage emergent chain-of-thought behaviors. Experimental results on synthetically generated mazes show that while a baseline model fails to navigate the maze, the SFT-trained model achieves 86\% accuracy, and further GRPO fine-tuning boosts accuracy to 93\%. Qualitative analyses reveal that GRPO fosters more robust and self-corrective reasoning, highlighting the potential of our approach to bridge the gap between language models and visual spatial tasks. These findings offer promising implications for applications in robotics, autonomous navigation, and other domains that require integrated visual and sequential reasoning.
\end{abstract}

\input{sections/1_introduction.tex}
\input{sections/2_related_work.tex}
\input{sections/3_methodology.tex}
\input{sections/4_experiments.tex}
\input{sections/5_discussion.tex}
\input{sections/6_conclusion.tex}

\bibliographystyle{plainnat} 
\bibliography{bibliography} 


\clearpage
\appendix
\onecolumn
\input{sections/appendices.tex}


\end{document}

%% file: sections/1_introduction.tex
\section{Introduction}
\label{sec:introduction}

The ability to reason about visual information, particularly in spatial contexts, is a hallmark of intelligent systems. From navigating physical environments to interpreting complex diagrams, visual spatial reasoning is crucial for a wide range of tasks. While Large Language Models (LLMs) have achieved impressive performance in natural language processing and code generation, their capacity for genuine visual reasoning, especially spatial understanding and sequential decision-making in visual environments, remains a significant open question \citep{zhang2024visionlanguagemodelsvisiontasks, ma2024surveyvisionlanguageactionmodelsembodied}. Current Vision-Language Models (VLMs) often excel at pattern recognition and object identification but may struggle with tasks requiring deeper spatial inference and step-by-step planning

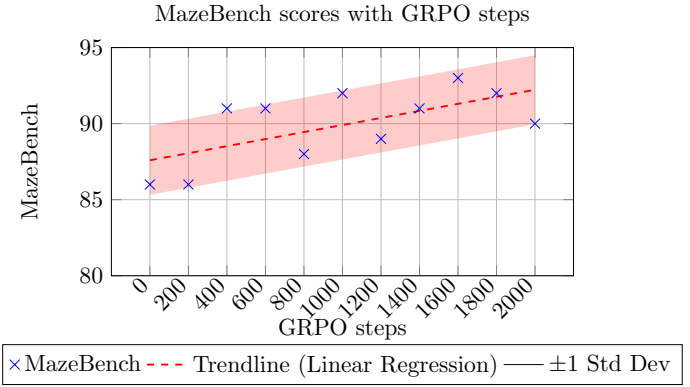
\begin{figure}[htbp]
  \centering
  \resizebox{1.05\linewidth}{!}{\input{figures/mazebench.tex}}
  \caption{MazeBench scores over GRPO steps with a linear regression trendline and its $\pm1$ standard deviation bounds.}
  \label{fig:maze_plot}
\end{figure}

in visual domains \citep{ma2024surveyvisionlanguageactionmodelsembodied}. Bridging this gap and endowing standard LLMs with robust visual reasoning capabilities is a critical step towards more versatile and human-like AI.

In this paper, we address the challenge of teaching visual spatial reasoning to a standard LLM, focusing on the task of maze navigation. We hypothesize that by providing an LLM with a tokenized \textit{visual} representation of a maze, we can train it to learn step-by-step movement commands to navigate from a designated origin to a target. The core of our approach lies in a two-stage training framework. First, we employ Supervised Fine-Tuning (SFT) to equip the LLM with the foundational skill of predicting movement tokens based on the visual maze input. Subsequently, we apply Group Relative Policy Optimization (GRPO), drawing inspiration from recent advancements in reinforcement learning for reasoning in LLMs, such as DeepSeek-R1 \citep{Guo2025DeepSeekR1}. DeepSeek-R1 demonstrated that Reinforcement Learning (RL) can elicit emergent reasoning behaviors, including chain-of-thought, even without prior SFT. We adapt and extend these RL strategies, combined with carefully designed reward functions, to refine our model's visual reasoning process for maze navigation.

To systematically evaluate LLM's ability to solve maze, we introduce MazeBench—a comprehensive benchmark on solving maze. MazeBench provides a controlled yet diverse environment that spans a range of maze sizes and complexities. By evaluating our model on MazeBench, we can rigorously measure both its maze-solving accuracy and the sophistication of its emergent reasoning behavior.

Our key contributions are as follows:

\begin{itemize}
    \item We present a novel training framework that combines Supervised Fine-Tuning and Group Relative Policy Optimization to enhance \textit{visual} reasoning in standard LLMs, specifically for spatial tasks.
    \item We empirically demonstrate that this framework, using a tokenized visual maze representation, enables an LLM to achieve improved maze navigation accuracy and exhibit emergent chain-of-thought reasoning in generating movement sequences.
    \item We provide a detailed analysis of the design and impact of reward functions within the GRPO stage, highlighting their crucial role in shaping the model's visual reasoning performance.
    \item We draw comparisons with insights from state-of-the-art reasoning models like DeepSeek-R1, both in terms of methodology and observed emergent behaviors, positioning our work within the context of current advancements in LLM reasoning.
    \item We present MazeBench, a benchmark for visual maze navigation that captures a wide spectrum of spatial challenges.
\end{itemize}

%% file: figures/mazebench.tex
\begin{tikzpicture}
  \begin{axis}[
    width=8.5cm,
    height=5cm,
    xlabel={GRPO steps},
    ylabel={MazeBench},
    title={MazeBench scores with GRPO steps},
    xtick={0,1,2,3,4,5,6,7,8,9,10},
    xticklabels={0,200,400,600,800,1000,1200,1400,1600,1800,2000},
    xticklabel style={rotate=45, anchor=east},
    ymin=80, ymax=95,
    grid=both,
    legend style={at={(0.5,-0.3)}, anchor=north,legend columns=-1}
  ]
    \addplot[only marks, mark=x, blue, mark size=3] coordinates {
      (0,86) (1,86) (2,91) (3,91) (4,88) (5,92) (6,89) (7,91) (8,93) (9,92) (10,90)
    };
    \addlegendentry{MazeBench}
    
    \addplot[domain=0:10, samples=11, red, dashed, thick, name path=trend] {0.4636*x + 87.5909};
    \addlegendentry{Trendline (Linear Regression)}
    
    \addplot[domain=0:10, samples=11, draw=none, name path=upper] {0.4636*x + 87.5909 + 2.27};
    \addplot[domain=0:10, samples=11, draw=none, name path=lower] {0.4636*x + 87.5909 - 2.27};
    \addplot[red, opacity=0.2] fill between[of=upper and lower];
    \addlegendentry{$\pm 1$ Std Dev}
    
  \end{axis}
\end{tikzpicture}

%% file: sections/2_related_work.tex
\section{Related Work}
\label{sec:related_work}

\subsection{Chain-of-Thought Reasoning in Language Models}

Chain-of-Thought (CoT) prompting has emerged as a powerful technique to elicit complex reasoning from Large Language Models \citep{Wei2022ChainofThought}. By prompting LLMs to "think step by step," CoT encourages the generation of intermediate reasoning steps, leading to improved performance on tasks requiring multi-step inference. Prior research, including \citet{Wei2022ChainofThought}, \cite{wei2023chainofthoughtpromptingelicitsreasoning}, \cite{wang-etal-2023-towards} prompting significantly enhances LLM performance on arithmetic, commonsense reasoning, and symbolic reasoning tasks. Our work builds upon the concept of CoT reasoning, aiming to induce a similar step-by-step thought process in LLMs, but within the domain of visual spatial reasoning for maze navigation.

\subsection{Supervised Fine-Tuning for Visual and Spatial Tasks}

Supervised Fine-Tuning (SFT) is a widely adopted technique for adapting pre-trained LLMs to specific downstream tasks \citep{wei2022finetunedlanguagemodelszeroshot}. By training on task-specific datasets, SFT allows LLMs to acquire specialized skills and improve performance in targeted domains.  \citet{jiang2024supervisedfinetuningturnimproves} recently highlighted the effectiveness of SFT in enhancing visual foundation models, demonstrating its utility in visual tasks.  In our research, we leverage SFT as the initial stage of our training pipeline, using it to equip the LLM with the basic capability of processing tokenized visual maze inputs and predicting movement tokens.  This SFT phase serves as a crucial foundation upon which we build more sophisticated reasoning through reinforcement learning.

\subsection{Reinforcement Learning and GRPO for Reasoning and Reward Shaping}

Reinforcement Learning from Human Feedback (RLHF) and its variants have demonstrated significant efficacy in aligning Large Language Models (LLMs) with human preferences and enhancing their reasoning capabilities. However, RLHF faces substantial scalability challenges due to its resource-intensive nature and reliance on human feedback data. As an alternative approach, recent methodologies like Group Relative Policy Optimization (GRPO)\citep{kwon2023rewarddesignlanguagemodels} and Self-Play fIne-tuNing (SPIN) leverage self-play mechanisms, where models autonomously generate training signals and iteratively improve through self-competition \cite{chen2024selfplayfinetuningconvertsweak}. These self-play approaches show promise in achieving human-level performance without the need for extensive human feedback, potentially offering a more scalable solution to the alignment challenge. GRPO, as described by \citet{shao2024deepseekmathpushinglimitsmathematical} and implemented in DeepSeek-R1 \citep{Guo2025DeepSeekR1}, offers a computationally efficient approach to reinforcement learning by estimating advantages based on group scores, eliminating the need for a separate critic network.  Reward function design is paramount in RLHF and GRPO, as it directly guides the model's learning process.  Carefully crafted reward functions can incentivize desired behaviors and shape the model's policy towards optimal performance.  Our work draws inspiration from the reward shaping strategies used in DeepSeek-R1 and adapts them to the context of visual maze navigation, designing reward components to encourage accuracy, valid movement sequences, and proper output formatting.

\subsection{DeepSeek-R1 and Emergent Reasoning through RL}

The DeepSeek-R1 model \citep{Guo2025DeepSeekR1} represents a significant advancement in using reinforcement learning to elicit sophisticated reasoning capabilities in LLMs.  A key finding of DeepSeek-R1 is the demonstration that pure RL, specifically GRPO, can lead to the \textit{emergent} development of chain-of-thought reasoning and even "aha moments," where the model re-evaluates previous steps and corrects its reasoning process.  Furthermore, DeepSeek-R1 highlights the benefits of a multi-stage training pipeline, combining initial RL training with subsequent supervised fine-tuning to refine language coherence and readability.  We directly adapt the GRPO optimization strategy and multi-stage training insights from DeepSeek-R1 to our visual maze navigation task.  We hypothesize that similar RL techniques can drive the emergence of visual spatial reasoning in standard LLMs, enabling them to solve mazes through a step-by-step, self-corrective process.

\subsection{Visual Reasoning and Maze Solving in AI}

Maze solving has long been a benchmark task in Artificial Intelligence, serving as a testbed for various problem-solving and search algorithms \citep{MazeSolverRobotScholarWorks}. Traditional approaches include graph search algorithms like Depth-First Search, Breadth-First Search, and A* \citep{PathfindingAlgorithmsRedBlobGames}. More recently, AI techniques, particularly reinforcement learning and neural networks, have been applied to maze navigation \citep{DeepRLMazeSolvingSamyzaf}.
While Chain-of-Thought (CoT) prompting has significantly enhanced complex reasoning capabilities in Large Language Models (LLMs) and Multimodal LLMs, it shows limitations in complex spatial reasoning tasks. Recent work by Microsoft introduces Multimodal Visualization-of-Thought (MVoT), which enables models to generate visual representations during their reasoning process, similar to human visual thinking \citep{li2025imaginereasoningspacemultimodal}. This breakthrough demonstrates the potential of combining verbal and visual reasoning in AI systems.

Our research builds upon these advances, focusing on teaching visual maze reasoning to standard language models through a tokenized visual representation and a combination of SFT and GRPO. This approach differs from traditional maze solvers by leveraging the inherent reasoning capabilities of LLMs and adapting them to process and reason about visual spatial information. Furthermore, research in neural-symbolic visual reasoning \citep{NeuralSymbolicVisualReasoningArxiv} explores combining neural networks with symbolic AI for visual tasks, offering a complementary perspective on integrating reasoning and visual processing.

%% file: sections/3_methodology.tex
\section{Methodology}
\label{sec:methodology}

\subsection{Tokenized Visual Maze Representation}{\label{sec:3.1}}
\label{subsec:tokenized_maze}
\begin{figure}
    \centering
    \includegraphics[width=0.5\linewidth]{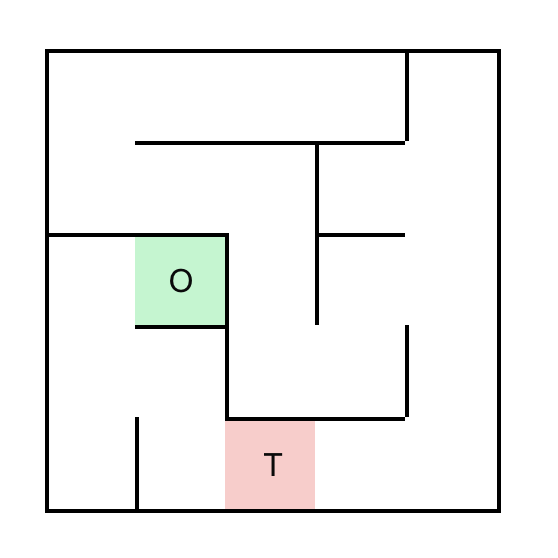}
    \caption{Visual of the Example Maze}
    \label{fig:example_maze}
\end{figure}
To enable the LLM to process maze information visually, we designed a tokenized input format that represents the maze grid, walls, origin, and target locations. Each cell in the maze is represented by a coordinate token \texttt{<|row-col|>}, e.g., \texttt{<|0-0|>} for the top-left cell. Wall information for each cell is encoded using tokens such as \texttt{<|no\_wall|>, <|up\_wall|>, <|up\_down\_wall|>, <|up\_down\_left\_right\_wall|>, ...}. The origin and target locations are marked with \texttt{<|origin|>} and \texttt{<|target|>} tokens, respectively. Empty spaces within the maze representation are filled with \texttt{<|blank|>} tokens for consistent grid structure. This tokenization scheme provides a visual representation by explicitly encoding the spatial relationships between cells and the presence of walls, allowing the LLM to ``see'' the maze structure in a symbolic, tokenized form.

\medskip
\noindent \textbf{Example Maze Tokenization \ref{fig:example_maze}:}

\begin{lstlisting}
<|0-0|><|up_left_wall|><|blank|><|0-1|><|up_down_wall|><|blank|><|0-2|><|up_down_wall|><|blank|><|0-3|><|up_down_right_wall|><|blank|><|0-4|><|up_left_right_wall|><|blank|>
<|1-0|><|down_left_wall|><|blank|><|1-1|><|up_down_wall|><|blank|><|1-2|><|up_right_wall|><|blank|><|1-3|><|up_down_left_wall|><|blank|><|1-4|><|right_wall|><|blank|>
<|2-0|><|up_left_wall|><|blank|><|2-1|><|up_down_right_wall|><|origin|><|2-2|><|left_right_wall|><|blank|><|2-3|><|up_left_wall|><|blank|><|2-4|><|right_wall|><|blank|>
<|3-0|><|left_wall|><|blank|><|3-1|><|up_right_wall|><|blank|><|3-2|><|down_left_wall|><|blank|><|3-3|><|down_right_wall|><|blank|><|3-4|><|left_right_wall|><|blank|>
<|4-0|><|down_left_right_wall|><|blank|><|4-1|><|down_left_wall|><|blank|><|4-2|><|up_down_wall|><|target|><|4-3|><|up_down_wall|><|blank|><|4-4|><|down_right_wall|><|blank|>
\end{lstlisting}

\subsection{Baseline Models}
\label{subsec:baseline_model}

To establish performance benchmarks for our approach, we employed three distinct baseline models, leveraging the DeepSeek-R1 \citep{Guo2025DeepSeekR1} Distill-Qwen family of language models. We evaluate two distilled models: DeepSeek-R1-Distill-Qwen-7B and DeepSeek-R1-Distill-Qwen-1.5B. Additionally, a Direct Prediction baseline was established using a Supervised Fine-Tuning (SFT) approach on the DeepSeek-R1-Distill-Qwen-1.5B architecture. This model was trained to directly predict the complete sequence of movement tokens representing the solution path through a given maze. The training objective was the minimization of cross-entropy loss between the predicted token sequence and the ground truth solution. This baseline assesses the performance of a standard language model trained to generate complete solutions without intermediate reasoning steps or reinforcement learning techniques. 

We include these three baselines to provide a comprehensive comparison, examining the influence of model size (7B vs. 1.5B) and the effectiveness of direct prediction versus our proposed step-by-step and reinforcement learning approaches. The subsequent sections will primarily focus on the customized direct prediction model and its enhancements through SFT for step-by-step reasoning and GRPO.
\subsection{Supervised Fine-Tuning (SFT) for Step-by-Step Reasoning}
\label{subsec:sft}

\begin{figure}
    \centering
    \includegraphics[width=0.9\linewidth]{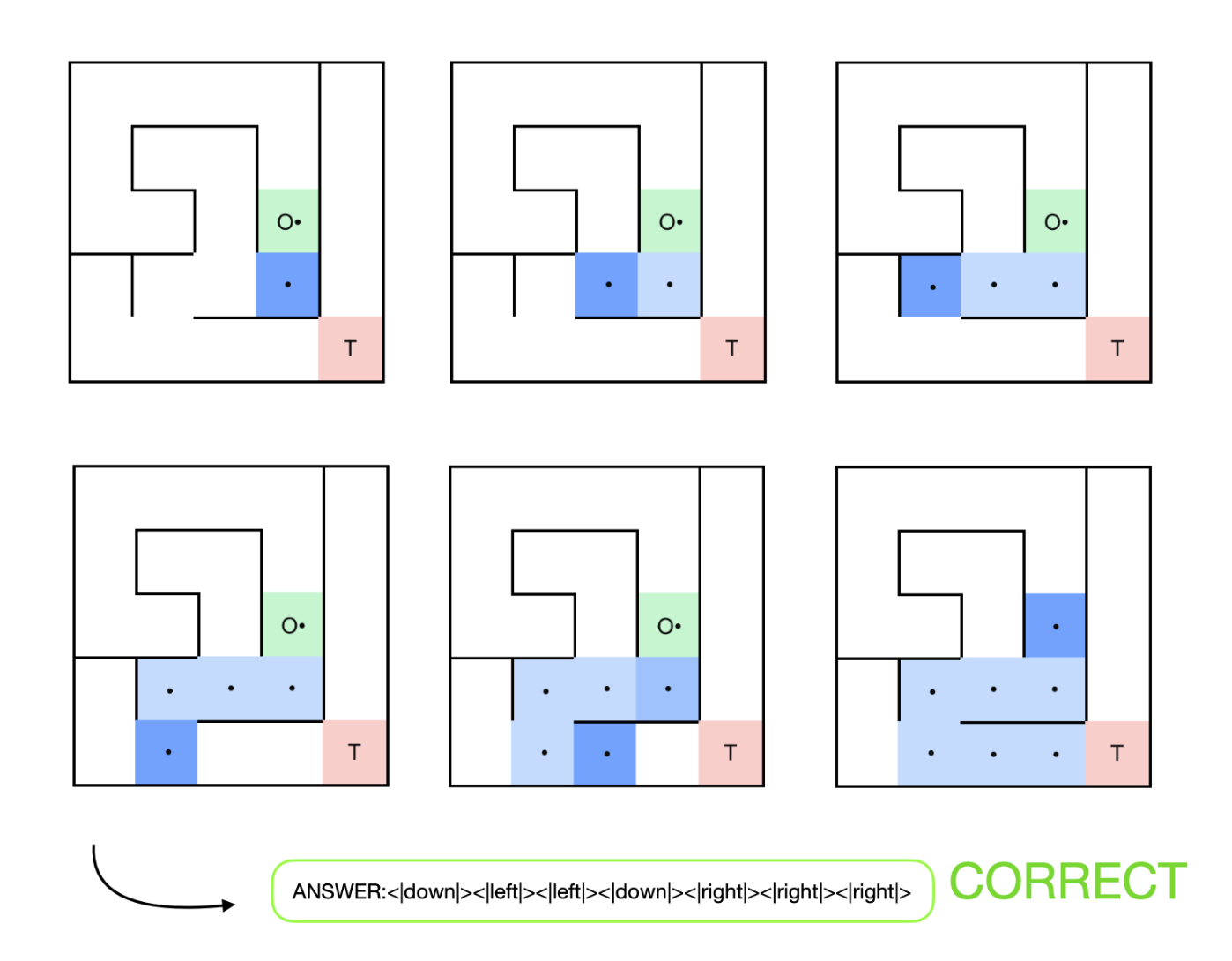}
    \caption{Visualization of AlphaMaze's step-by-step reasoning process while solving a maze.}
    \label{fig:visual-reasoning}
\end{figure}

For the SFT stage, we curated a training dataset. Mazes were synthetically generated with fixed sizes (5x5) and varied complexity level. The Qwen 1.5B SFT model was then trained on this dataset. The training objective was to predict the \textit{next} movement token at each step, conditioned on the maze input and the preceding movement tokens in the sequence as visually illustrated in Figure \ref{fig:visual-reasoning}. This step-by-step prediction approach was designed to encourage the model to learn sequential reasoning for maze navigation.

\subsection{Group Relative Policy Optimization (GRPO) for Enhanced Reasoning}
\label{subsec:grpo}

Following SFT, we applied Group Relative Policy Optimization (GRPO) to further enhance the model's maze-solving capabilities and encourage more robust reasoning. The GRPO training utilized a smaller set of data than SFT state. We designed a reward function 3 components.

\textbf{Correctness Reward (+0.2 per solution step):} This reward is scaled according to the number of steps in the maze solution. Each valid movement step adds 0.2 points to the total score. For example, a solution requiring 4 steps earns a reward of \(0.2 \times 4 = 0.8\) points, incentivizing both accuracy and efficiency in navigation.

\textbf{Integrity Reward (+0.5):} This reward is given for each valid movement token (\texttt{<|up|>}, \texttt{<|down|>}, \texttt{<|left|>}, \texttt{<|right|>}) in the predicted sequence, encouraging the generation of meaningful and valid movement steps.

\textbf{Thinking Reward (+0.25):} This reward is given for correctly using the \texttt{<think>} tag in the output, ensuring completeness and consistency in the reasoning format.

These reward components were weighted to prioritize correctness while also encouraging valid movement sequences and proper reasoning formatting with \texttt{<think>} tag. We adapted the Group Relative Policy Optimization (GRPO) algorithm, as employed in DeepSeek-R1 \citep{Guo2025DeepSeekR1}, to perform reinforcement learning. GRPO estimates advantages based on relative group scores, offering computational efficiency compared to critic-based methods.

\subsection{Training Procedure and Pipeline}
\label{subsec:training_pipeline}

Our training pipeline consisted of two stages. First, \textbf{Supervised Fine-Tuning (SFT)} was performed on the Qwen 1.5B model using a curated maze dataset for 10 epochs to learn step-by-step movement prediction for maze navigation. This phase established a strong initial policy, ensuring that the model could effectively interpret and respond to sequential movement tasks.  

Following SFT, \textbf{Group Relative Policy Optimization (GRPO)} was applied to refine the model’s performance. The SFT-trained model was further fine-tuned using LoRA \cite{hu2021loralowrankadaptationlarge} with the GRPO method, implemented in Unsloth \cite{unsloth} with VLLM \cite{kwon2023efficient} for efficient inference. A carefully designed reward function guided the optimization process, and model checkpoints were saved every 200 steps to track improvements.  

This two-stage pipeline mirrors the multi-stage training approach employed in DeepSeek-R1 \citep{Guo2025DeepSeekR1}, where initial RL training is followed by supervised fine-tuning for refinement. In our case, SFT provided a robust starting point for reinforcement learning (RL), allowing GRPO to focus on refining reasoning capabilities and enhancing task-specific performance.

%% file: sections/4_experiments.tex
\section{Experiments and Results}
\label{sec:experiments}

\subsection{Dataset Details}
\label{subsec:dataset_details}
The dataset is constructed through a multi-stage process involving generation, refinement, and augmentation. The process begins with the creation of a large initial pool of 530,000 synthetic mazes. These mazes are generated using the \textbf{maze-dataset} framework \cite{ivanitskiy2023configurablelibrarygeneratingmanipulating}, which employs a randomized depth-first search algorithm. This algorithm ensures that every generated maze has a guaranteed solution path connecting the designated origin and target locations. Further details of algorithm used can be found at Appendix \ref{alg:main}. All mazes within the dataset have a fixed size of 5x5 grids. From this extensive initial pool, a subset of 30,000 mazes is randomly selected and reserved as a held-out test set. This separation guarantees that the training and evaluation data are entirely distinct, preventing data leakage and enabling a robust assessment of model generalization.

The remaining 500,000 mazes form the basis for the various training datasets used in this work. This pool of mazes undergoes a multi-stage processing and augmentation procedure to create datasets specifically tailored for different training objectives.

\textit{\textbf{Reset Dataset Creation:}} A significant portion of the training data focuses on teaching the model to recover from errors. To this end, a "reset" dataset is created. This dataset is generated by algorithmically producing \textbf{incorrect} solution paths for the mazes. These incorrect paths are designed to be plausible but ultimately unsuccessful, either leading to dead ends or deviating from the correct solution. Importantly, they adhere to constraints: they do not revisit locations already visited within the incorrect attempt, and they avoid portions of the known correct solution path.

Associated with each incorrect path, a textual "RESET" message is generated, simulating the feedback a system might provide upon encountering an error.  The content of this message depends on whether the incorrect path terminates at a dead end (three surrounding walls) or simply deviates from the wrong route.  These incorrect paths, along with their reset messages and the \textbf{correct} solution's Chain-of-Thought (COT) reasoning, are combined.  This process results in approximately 400,000 training examples where the model is presented with scenarios of initial failure(s) followed by a successful attempt after a "reset."  The intent is to train the model to recognize incorrect trajectories and adapt its strategy. Illustrative examples of reset samples are provided in Appendix.

 \textit{\textbf{SFT Training Data Construction:}} The final Supervised Fine-Tuning (SFT) training dataset is a balanced combination of "straight success" examples and "retry" examples:
\begin{itemize}
    \item \textit{Straight Success Data} (250,000 mazes):  This portion consists of mazes where the model is expected to generate the correct solution path on the first attempt, without any resets.
    \item \textit{Retry Data} (250,000 mazes): This portion is drawn from the "reset" dataset described above, providing examples where the model learns from incorrect attempts and subsequent resets. This combined 500,000-maze SFT set, encompassing success and error recovery, enables robust learning.
\end{itemize}
 \textit{\textbf{GRPO Training Data:} } The remaining 150,000 mazes from the original "straight success" data pool are used to create a dataset for GRPO stage.

\subsection{MazeBench}

To rigorously evaluate the spatial reasoning and planning capabilities of large language models (LLMs), we introduce MazeBench, a novel benchmark consisting of a curated collection of 100 maze-solving challenges. While existing benchmarks often assess logical reasoning or commonsense knowledge, MazeBench specifically targets the ability of LLMs to understand spatial relationships, plan multi-step paths, and execute sequential actions within a constrained environment. This capacity is crucial for applications ranging from robotics and navigation to game playing and virtual agent control.

MazeBench is a collection of 100 unique mazes, randomly selected from a larger test set containing 30,000 mazes. It is designed to evaluate the performance of large language models (LLMs) by categorizing mazes into different difficulty levels. Each maze requires the model to determine an optimal path from the starting point to the target, with difficulty primarily based on the number of steps needed to reach the goal.

The benchmark is structured into three levels: Easy, Medium, and Hard, ensuring a progressive assessment of an LLM’s pathfinding and problem-solving abilities. The components are described in Table \ref{tab:maze-config}.

\begin{table}[htbp]
    \centering
    \caption{Maze Configuration by Difficulty Level}
    \label{tab:maze-config}
    \begin{tabular}{|l|c|c|}
        \hline
        Category & Number of Mazes & Steps \\
        \hline
        Easy   & 50 & 1 -- 4 \\
        Medium & 40 & 5 -- 8 \\
        Hard   & 10 & 9 -- 13 \\
        \hline
        \textbf{Total} & \textbf{100} & \textbf{1 -- 13} \\
        \hline
    \end{tabular}
\end{table}

\textbf{The Easy category} consists of 50 mazes, each requiring between 1 and 4 steps to solve. These simpler mazes establish a baseline for evaluating fundamental navigation skills.

\textbf{The Medium category} includes 40 mazes that demand solution paths of 5 to 8 steps. These mazes introduce a higher level of complexity, requiring more advanced planning and spatial reasoning. Successfully solving them indicates an LLM’s ability to manage moderately intricate environments.

\textbf{The Hard category} comprises 10 mazes, each necessitating 9 to 13 steps to reach the target. These mazes present the greatest challenge, testing an LLM’s capacity to handle long-range dependencies and navigate complex spatial structures. Performance on this level reflects the model’s ability to process and reason over extended solution paths.

As mentioned previously, the mazes are presented to the LLM in a tokenized input format; the full details of this representation, including examples, are provided in Section~\ref{sec:3.1}. The LLM is expected to produce sequence of movement tokens. During evaluation, we will parse the LLM's output to extract these tokens. The order of these tokens is crucial. The presence of extraneous characters, whitespace, or other tokens will not automatically invalidate the solution, provided that the correct sequence of movement tokens can be extracted. A solution is considered incorrect if the extracted sequence of movement tokens does not lead to the target or leads to an invalid state (e.g., attempting to move into a wall) is considered incorrect. The evaluation metric is the success rate: the percentage of mazes solved correctly.

\subsection{Quantitative Results}
\label{subsec:experimental_results}

\subsubsection{Model Performance on MazeBench}
\label{subsubsec:accuracy}
As shown in Table~\ref{tab:accuracy}, the baseline model, trained for direct path prediction without explicit reasoning, achieved 0\% accuracy on MazeBench. This highlights the necessity of step-by-step reasoning for the task. The SFT-only model reached a baseline of 86.0\%, demonstrating the effectiveness of supervised fine-tuning for learning step-by-step maze navigation. Further enhancement with GRPO led to significant improvement, reaching 93.0\% after 1600 steps of GRPO training.

\begin{table}[htbp]
    \centering
    \caption{Maze Solving Accuracy on MazeBench}
    \label{tab:accuracy}
    \begin{tabular}{|l|c|c|c|}
        \hline
        Model & SFT & GRPO & Score \\
        \hline
        Baseline-1.5B & \ding{55} & \ding{55} & 0.0 \\   
        Baseline-7B & \ding{55} & \ding{55} & 0.0 \\
        Baseline-1.5B (SFT) & \checkmark & \ding{55} & 0.0 \\
        AlphaMaze-SFT & \checkmark & \ding{55} & 86.0 \\
        AlphaMaze & \checkmark & \checkmark & \textbf{93.0} \\
        \hline
    \end{tabular}
\end{table}

\subsubsection{Model Evolution During GRPO}
\label{subsubsec:training_dynamics}

Figure~\ref{fig:maze_plot} displays the MazeBench scores (blue crosses) over GRPO steps along with a linear regression trendline (red dashed line) and its $\pm1$ standard deviation bounds. The steady increase in the trendline indicates that GRPO effectively guides the model towards improved maze-solving policies.

\subsection{Qualitative Results}
\label{subsec:qualitative_results}

Qualitative analysis of model outputs revealed notable differences in reasoning behavior. The baseline model often produced nonsensical or incomplete movement sequences, frequently failing to reach the target and exhibiting "hallucinations" by predicting movements invalid within the maze structure. The \textbf{AlphaMaze-SFT} model demonstrated improved coherence and step-by-step progression, but still struggled with longer or more complex mazes, sometimes becoming trapped in loops or making incorrect turns in later stages of the solution path.

In contrast, the \textbf{AlphaMaze-SFT+GRPO} model exhibited the most sophisticated reasoning. In many instances, emergent chain-of-thought patterns were observed, with AlphaMaze (two-stage) appearing to explicitly consider wall constraints and spatial relationships at each step before predicting the next movement. Furthermore, outputs occasionally displayed instances reminiscent of the "aha moments" reported in prior work on DeepSeek-R1. For example, in some complex mazes, AlphaMaze (two-stage) would initially begin along one path, then appear to "re-evaluate" its trajectory mid-sequence, correcting its course to find a more efficient or correct solution. Error analysis indicated that AlphaMaze (two-stage) made fewer invalid moves and was more robust to long-context reasoning challenges compared to the AlphaMaze-SFT model. However, limitations remained, particularly in mazes requiring backtracking or complex spatial planning beyond the immediate next step.

%% file: sections/5_discussion.tex
\section{Discussion}
\label{sec:discussion}

\subsection{Analysis of GRPO's Impact on Visual Maze Reasoning}
\label{subsec:grpo_impact_analysis}

Our results clearly demonstrate the incremental benefit of Group Relative Policy Optimization (GRPO) in enhancing visual maze reasoning within Large Language Models.  While Supervised Fine-Tuning (SFT) establishes a strong foundation, enabling the model to achieve a \textbf{86\%} accuracy on MazeBench, the application of GRPO further elevates performance to \textbf{93\%} after 1600 training steps. This improvement, albeit seemingly modest in percentage points, is significant considering the already strong baseline established by SFT. It suggests that GRPO is effectively refining the model's policy, leading to more robust and accurate maze navigation.

The qualitative analysis provides further insight into the nature of this improvement.  The \textbf{AlphaMaze-SFT+GRPO} model exhibited more pronounced chain-of-thought reasoning patterns and instances of self-correction, indicating that GRPO is not merely fine-tuning the existing SFT policy, but rather encouraging more sophisticated reasoning processes. The reward function, designed to incentivize correctness, valid movements, and structured output, likely plays a crucial role in shaping this behavior. By rewarding successful navigation and penalizing invalid steps, GRPO encourages the model to learn more deliberate and considered movement strategies.

\subsection{Comparison with DeepSeek-R1 and RL for Reasoning}
\label{subsec:comparison_deepseek_r1}

It is important to note that the base DeepSeek-R1 model, when operating with an extremely long context window, demonstrates emergent visual reasoning capabilities. However, our experiments reveal that the distilled variants (DeepSeek-R1 Distill-Qwen models) do not carry over these spatial reasoning abilities, as evidenced by their \textbf{0\%} accuracy on MazeBench. This suggests that the distillation process into Qwen or other smaller models is insufficient to preserve the emergent ability of visual spatial reasoning observed in the base model.

In contrast, our two-stage training approach—combining Supervised Fine-Tuning (SFT) to establish foundational step-by-step reasoning with Group Relative Policy Optimization (GRPO) for further refinement—effectively equips the distilled model with robust visual maze-solving skills. Even with only 1600 GRPO steps, the model achieves a notable improvement, reaching \textbf{93\%} accuracy and exhibiting clear chain-of-thought behaviors along with self-correction during navigation.

These findings underscore the necessity of specialized training to recover or enhance spatial reasoning in distilled models, highlighting that while the base DeepSeek-R1 model is capable of visual reasoning with sufficient context, additional training stages are crucial to maintain or induce this capability in smaller, distilled variants.

\subsection{Limitations}
\label{subsec:limitations}

Despite the encouraging results, our study is not without limitations.  Firstly, the performance gain from GRPO, while statistically significant, is small (7\% accuracy improvement in our reported experiment).  Further investigation is needed to explore whether more extensive GRPO training, or modifications to the reward function, could lead to more substantial performance gains.  It is possible that the current reward function, while effective, could be further optimized to better incentivize more complex reasoning strategies, such as backtracking or more proactive exploration of alternative paths.

Secondly, our evaluation, while including qualitative analysis, is primarily based on maze-solving accuracy.  This metric, while important, provides a somewhat limited view of the model's reasoning capabilities.  Future work could benefit from more nuanced evaluation metrics that assess the efficiency of the generated paths, the robustness of the model to maze complexity variations, and the interpretability of the model's internal reasoning process.  Furthermore, while we observed qualitative signs of chain-of-thought reasoning, a more rigorous analysis, perhaps using techniques from interpretability research, is needed to definitively characterize the nature and depth of the model's reasoning process.

Finally, our experiments are limited to synthetically generated mazes.  While these mazes were designed to vary in size and complexity, they may not fully capture the intricacies and variability of real-world visual spatial reasoning tasks.  Future research should explore the generalizability of our approach to more diverse and ecologically valid visual environments and tasks.

%% file: sections/6_conclusion.tex
\section{Conclusion}
\label{sec:conclusion}

This paper introduced AlphaMaze, a novel approach to enhance Large Language Models' spatial intelligence, focusing on maze navigation. We demonstrated the efficacy of a two-stage training framework, leveraging Supervised Fine-Tuning (SFT) followed by Group Relative Policy Optimization (GRPO). While initial pre-trained LLMs exhibited 0\% accuracy on MazeBench, highlighting the need for task-specific adaptation, our approach successfully imbued a distilled LLM with robust spatial reasoning capabilities. SFT provided a crucial foundation by teaching step-by-step movement prediction from tokenized maze inputs, reaching 86\% accuracy. This underscores the importance of structured input and targeted training for LLMs to effectively engage with visual spatial information.

Crucially, we adapted and applied the two-stage training methodology pioneered by DeepSeek-R1, demonstrating its generalizability beyond language-centric reasoning tasks. Following SFT, GRPO fine-tuning further elevated performance to 93\% on MazeBench after 1600 training steps, showcasing the power of reinforcement learning to refine reasoning processes in a novel domain. Qualitative analysis revealed that GRPO fostered more sophisticated and self-corrective reasoning strategies, including emergent chain-of-thought patterns, mirroring observations in DeepSeek-R1 and suggesting a common mechanism for enhanced reasoning through RL.

Our work contributes to the broader effort of expanding LLMs' reasoning abilities beyond natural language, demonstrating the potential of a two-stage approach for visually grounded tasks. The success of GRPO, inspired by DeepSeek-R1's advancements in language reasoning, highlights the transferability of these techniques to spatial domains. This suggests that carefully designed reinforcement learning, following an initial phase of supervised task learning, can be a powerful method to unlock and refine sophisticated reasoning capabilities in LLMs across diverse problem spaces. The implications extend beyond maze navigation to a wide array of applications demanding spatial understanding and sequential decision-making.

Future research will focus on further validating this two-stage GRPO approach across various reasoning domains beyond spatial tasks, exploring its potential to enhance LLMs' capabilities in areas such as symbolic reasoning, logical deduction, and planning. Investigating the optimal configurations of SFT and GRPO stages, diversifying training data to encompass richer and more complex reasoning scenarios, and developing more refined reward functions tailored to different reasoning challenges are critical next steps. By pursuing these directions, we aim to establish the broader applicability of this two-stage training paradigm for imbuing standard LLMs with robust and versatile reasoning abilities, paving the way for more capable and generally intelligent language models.

%% file: sections/appendices.tex
\appendix

\section{Algorithm}
This appendix details the algorithm used to generate the maze reasoning dataset with reset demonstrations. The algorithm processes a base dataset of maze navigation problems and augments it with demonstration of incorrect attempts followed by resets and correct solutions.

\begin{algorithm}
\caption{Maze Reasoning Reset Data Generation - Main Process}
\label{alg:main}
\begin{algorithmic}[1]
\REQUIRE Base dataset $D$ containing maze problems with:
    \STATE \quad - Adjacency list representation of $5 \times 5$ maze grid
    \STATE \quad - Origin and target coordinates
    \STATE \quad - Correct solution path
\ENSURE Augmented dataset with reset demonstrations

\STATE Initialize empty datasets $D_1$ and $D_2$

\FORALL{example $e \in D$}
    \STATE Extract adjacency list $A$, origin $O$, target $T$, and path $P$ from $e$
    \STATE Count walls $W$ around origin $O$
    \IF{$W = 1$}
        \STATE Add $e$ to $D_1$
        \STATE Call ProcessOrder1($e$) \COMMENT{See Algorithm \ref{alg:order1}}
    \ELSIF{$W = 2$}
        \STATE Add $e$ to $D_2$
        \STATE Call ProcessOrder2($e$) \COMMENT{See Algorithm \ref{alg:order2}}
    \ENDIF
\ENDFOR

\STATE Combine processed examples from $D_1$ and $D_2$ into the final dataset
\end{algorithmic}
\end{algorithm}

\begin{algorithm}
\caption{Order-1 Processing (1 wall at origin)}
\label{alg:order1}
\begin{algorithmic}[1]
\STATE \textbf{Procedure} ProcessOrder1(example)
    \STATE $WP \leftarrow \emptyset$ \COMMENT{Initialize wrong paths set}
    \FORALL{adjacent node $N$ to origin $O$}
        \IF{$N \notin$ correct path $P$}
            \FOR{$n\_steps$ from $\text{max\_n\_steps}$ down to $1$}
                \STATE Attempt to extend path from $N$ until a dead end or $n\_steps$ are reached.
                \IF{path length = $n\_steps$ or a dead end is reached}
                    \STATE $WP \leftarrow WP \cup \{\text{path}\}$
                    \STATE \textbf{break}
                \ENDIF
            \ENDFOR
        \ENDIF
    \ENDFOR
    \FORALL{path $p \in WP$}
        \STATE Generate chain-of-thought steps for path $p$.
        \STATE Add ``Heading in wrong direction'' message.
        \STATE Add RESET marker.
    \ENDFOR
    \STATE Append original correct solution (path $P$).
    \STATE Format as conversation pairs.
\STATE \textbf{End Procedure}
\end{algorithmic}
\end{algorithm}

\begin{algorithm}
\caption{Order-2 Processing (2 walls at origin)}
\label{alg:order2}
\begin{algorithmic}[1]
\STATE \textbf{Procedure} ProcessOrder2(example)
    \FOR{$n\_steps$ from $\text{max\_n\_steps}$ down to $1$}
        \STATE Generate wrong path $WP$ of length $n\_steps$ starting from $O$.
        \IF{a valid path $WP$ is found}
            \STATE Generate chain-of-thought for $WP$.
            \IF{$WP$ ends at a dead end (3 walls)}
                \STATE Add ``Hit a dead end'' message.
            \ELSE
                \STATE Add ``Heading in wrong direction'' message.
            \ENDIF
            \STATE Add RESET marker.
            \STATE \textbf{break}
        \ENDIF
    \ENDFOR
    \STATE Append original correct solution (path $P$).
    \STATE Format as conversation pairs.
\STATE \textbf{End Procedure}
\end{algorithmic}
\end{algorithm}

%% file: main.bbl
\begin{thebibliography}{20}
\providecommand{\natexlab}[1]{#1}
\providecommand{\url}[1]{\texttt{#1}}
\expandafter\ifx\csname urlstyle\endcsname\relax
  \providecommand{\doi}[1]{doi: #1}\else
  \providecommand{\doi}{doi: \begingroup \urlstyle{rm}\Url}\fi

\bibitem[Chen et~al.(2024)Chen, Deng, Yuan, Ji, and Gu]{chen2024selfplayfinetuningconvertsweak}
Zixiang Chen, Yihe Deng, Huizhuo Yuan, Kaixuan Ji, and Quanquan Gu.
\newblock Self-play fine-tuning converts weak language models to strong language models, 2024.
\newblock URL \url{https://arxiv.org/abs/2401.01335}.

\bibitem[Daniel~Han and team(2023)]{unsloth}
Michael~Han Daniel~Han and Unsloth team.
\newblock Unsloth, 2023.
\newblock URL \url{http://github.com/unslothai/unsloth}.

\bibitem[Guo et~al.(2025)Guo, Xu, Chen, Tang, and Others]{Guo2025DeepSeekR1}
Duyu Guo, Guoxin Xu, Yuchen Chen, Chen Tang, and Others.
\newblock Deepseek-r1: Incentivizing reasoning capability in llms via reinforcement learning.
\newblock \emph{arXiv preprint arXiv:2501.12948}, 2025.
\newblock URL \url{https://arxiv.org/abs/2501.12948}.

\bibitem[Hu et~al.(2021)Hu, Shen, Wallis, Allen-Zhu, Li, Wang, Wang, and Chen]{hu2021loralowrankadaptationlarge}
Edward~J. Hu, Yelong Shen, Phillip Wallis, Zeyuan Allen-Zhu, Yuanzhi Li, Shean Wang, Lu~Wang, and Weizhu Chen.
\newblock Lora: Low-rank adaptation of large language models, 2021.
\newblock URL \url{https://arxiv.org/abs/2106.09685}.

\bibitem[Ivanitskiy et~al.(2023)Ivanitskiy, Shah, Spies, Räuker, Valentine, Rager, Quirke, Mathwin, Corlouer, Behn, and Fung]{ivanitskiy2023configurablelibrarygeneratingmanipulating}
Michael~Igorevich Ivanitskiy, Rusheb Shah, Alex~F. Spies, Tilman Räuker, Dan Valentine, Can Rager, Lucia Quirke, Chris Mathwin, Guillaume Corlouer, Cecilia~Diniz Behn, and Samy~Wu Fung.
\newblock A configurable library for generating and manipulating maze datasets, 2023.
\newblock URL \url{https://arxiv.org/abs/2309.10498}.

\bibitem[Janamian and Alam(2023)]{MazeSolverRobotScholarWorks}
Saba Janamian and MD~Sahabul Alam.
\newblock Maze solver robot using a* algorithm, 2023.
\newblock URL \url{https://scholarworks.calstate.edu/concern/theses/0c483r787}.
\newblock ScholarWorks@CSUN.

\bibitem[Jiang et~al.(2024)Jiang, Ge, Ge, Shi, Yuan, and Shan]{jiang2024supervisedfinetuningturnimproves}
Xiaohu Jiang, Yixiao Ge, Yuying Ge, Dachuan Shi, Chun Yuan, and Ying Shan.
\newblock Supervised fine-tuning in turn improves visual foundation models, 2024.
\newblock URL \url{https://arxiv.org/abs/2401.10222}.

\bibitem[Kwon et~al.(2023{\natexlab{a}})Kwon, Xie, Bullard, and Sadigh]{kwon2023rewarddesignlanguagemodels}
Minae Kwon, Sang~Michael Xie, Kalesha Bullard, and Dorsa Sadigh.
\newblock Reward design with language models, 2023{\natexlab{a}}.
\newblock URL \url{https://arxiv.org/abs/2303.00001}.

\bibitem[Kwon et~al.(2023{\natexlab{b}})Kwon, Li, Zhuang, Sheng, Zheng, Yu, Gonzalez, Zhang, and Stoica]{kwon2023efficient}
Woosuk Kwon, Zhuohan Li, Siyuan Zhuang, Ying Sheng, Lianmin Zheng, Cody~Hao Yu, Joseph~E. Gonzalez, Hao Zhang, and Ion Stoica.
\newblock Efficient memory management for large language model serving with pagedattention.
\newblock In \emph{Proceedings of the ACM SIGOPS 29th Symposium on Operating Systems Principles}, 2023{\natexlab{b}}.

\bibitem[Lester(2014-2024)]{PathfindingAlgorithmsRedBlobGames}
Patrick Lester.
\newblock Pathfinding algorithms, 2014-2024.
\newblock URL \url{https://www.redblobgames.com/pathfinding/}.
\newblock Red Blob Games.

\bibitem[Li et~al.(2025)Li, Wu, Zhang, Xia, Mao, Dong, Vulić, and Wei]{li2025imaginereasoningspacemultimodal}
Chengzu Li, Wenshan Wu, Huanyu Zhang, Yan Xia, Shaoguang Mao, Li~Dong, Ivan Vulić, and Furu Wei.
\newblock Imagine while reasoning in space: Multimodal visualization-of-thought, 2025.
\newblock URL \url{https://arxiv.org/abs/2501.07542}.

\bibitem[Ma et~al.(2024)Ma, Song, Zhuang, Hao, and King]{ma2024surveyvisionlanguageactionmodelsembodied}
Yueen Ma, Zixing Song, Yuzheng Zhuang, Jianye Hao, and Irwin King.
\newblock A survey on vision-language-action models for embodied ai, 2024.
\newblock URL \url{https://arxiv.org/abs/2405.14093}.

\bibitem[Mao et~al.(2023)Mao, Gan, Zhang, and Others]{NeuralSymbolicVisualReasoningArxiv}
Jiajun Mao, Chuang Gan, Fan Zhang, and Others.
\newblock Neural-symbolic visual reasoning: A survey.
\newblock 2023.
\newblock URL \url{https://arxiv.org/abs/2302.07200}.

\bibitem[Shao et~al.(2024)Shao, Wang, Zhu, Xu, Song, Bi, Zhang, Zhang, Li, Wu, and Guo]{shao2024deepseekmathpushinglimitsmathematical}
Zhihong Shao, Peiyi Wang, Qihao Zhu, Runxin Xu, Junxiao Song, Xiao Bi, Haowei Zhang, Mingchuan Zhang, Y.~K. Li, Y.~Wu, and Daya Guo.
\newblock Deepseekmath: Pushing the limits of mathematical reasoning in open language models, 2024.
\newblock URL \url{https://arxiv.org/abs/2402.03300}.

\bibitem[Wang et~al.(2023)Wang, Min, Deng, Shen, Wu, Zettlemoyer, and Sun]{wang-etal-2023-towards}
Boshi Wang, Sewon Min, Xiang Deng, Jiaming Shen, You Wu, Luke Zettlemoyer, and Huan Sun.
\newblock Towards understanding chain-of-thought prompting: An empirical study of what matters.
\newblock In Anna Rogers, Jordan Boyd-Graber, and Naoaki Okazaki, editors, \emph{Proceedings of the 61st Annual Meeting of the Association for Computational Linguistics (Volume 1: Long Papers)}, pages 2717--2739, Toronto, Canada, July 2023. Association for Computational Linguistics.
\newblock \doi{10.18653/v1/2023.acl-long.153}.
\newblock URL \url{https://aclanthology.org/2023.acl-long.153/}.

\bibitem[Wei et~al.(2022{\natexlab{a}})Wei, Bosma, Zhao, Guu, Yu, Lester, Du, Dai, and Le]{wei2022finetunedlanguagemodelszeroshot}
Jason Wei, Maarten Bosma, Vincent~Y. Zhao, Kelvin Guu, Adams~Wei Yu, Brian Lester, Nan Du, Andrew~M. Dai, and Quoc~V. Le.
\newblock Finetuned language models are zero-shot learners, 2022{\natexlab{a}}.
\newblock URL \url{https://arxiv.org/abs/2109.01652}.

\bibitem[Wei et~al.(2022{\natexlab{b}})Wei, Zhou, Le, Zhou, Le, and Others]{Wei2022ChainofThought}
Jason Wei, Denny Zhou, Quoc Le, Denny Zhou, Quoc Le, and Others.
\newblock Chain-of-thought prompting elicits reasoning in large language models.
\newblock \emph{Advances in Neural Information Processing Systems}, 35:\penalty0 24824--24837, 2022{\natexlab{b}}.
\newblock URL \url{https://proceedings.neurips.cc/paper_files/paper/2022/hash/9d8fc0533c2250385321d99c6a3f2f2c-Abstract-Conference.html}.

\bibitem[Wei et~al.(2023)Wei, Wang, Schuurmans, Bosma, Ichter, Xia, Chi, Le, and Zhou]{wei2023chainofthoughtpromptingelicitsreasoning}
Jason Wei, Xuezhi Wang, Dale Schuurmans, Maarten Bosma, Brian Ichter, Fei Xia, Ed~Chi, Quoc Le, and Denny Zhou.
\newblock Chain-of-thought prompting elicits reasoning in large language models, 2023.
\newblock URL \url{https://arxiv.org/abs/2201.11903}.

\bibitem[Zafrany(2020)]{DeepRLMazeSolvingSamyzaf}
Samy Zafrany.
\newblock Deep reinforcement learning for maze solving, 2020.
\newblock URL \url{https://www.samyzaf.com/ML/rl/qmaze.html}.
\newblock samyzaf.com.

\bibitem[Zhang et~al.(2024)Zhang, Huang, Jin, and Lu]{zhang2024visionlanguagemodelsvisiontasks}
Jingyi Zhang, Jiaxing Huang, Sheng Jin, and Shijian Lu.
\newblock Vision-language models for vision tasks: A survey, 2024.
\newblock URL \url{https://arxiv.org/abs/2304.00685}.

\end{thebibliography}
